\documentclass[conference]{IEEEtran}
\usepackage{xcolor}
\usepackage{cite}
\ifCLASSINFOpdf
  \usepackage[pdftex]{graphicx}
  \graphicspath{{figures/}}
\else
\fi

\usepackage{amsmath}

\usepackage{array}

\ifCLASSOPTIONcompsoc
 \usepackage[caption=false,font=normalsize,labelfont=sf,textfont=sf]{subfig}
\else
 \usepackage[caption=false,font=footnotesize]{subfig}
\fi

\usepackage{fixltx2e}

\usepackage{url}
\usepackage{amsmath}
\usepackage{amssymb}
\begin{document}
\title{AMI-FML: A Privacy-Preserving Federated Machine Learning Framework for AMI}

\markboth{}%
{Shell \MakeLowercase{\textit{et al.}}: }
%


\author{\IEEEauthorblockN{Milan Biswal}
\IEEEauthorblockA{\textit{Computer Science Department} \\
\textit{New Mexico State University, USA}\\
milanb@nmsu.edu}
\and
\IEEEauthorblockN{Abu-Saleh Md Tayeen}
\IEEEauthorblockA{\textit{Computer Science Department} \\
\textit{New Mexico State University, USA}\\
tayeen@nmsu.edu}
\and
\IEEEauthorblockN{Satyajayant~Misra}
\IEEEauthorblockA{\textit{Computer Science Department} \\
\textit{New Mexico State University, USA}\\
misra@nmsu.edu}
}
\maketitle
\begin{abstract}
Machine learning (ML) based smart meter data analytics is very promising for energy management and demand response applications in the advanced metering infrastructure (AMI). A key challenge in developing distributed ML applications for AMI is to preserve user privacy while allowing active end-users participation. This paper addresses this challenge and proposes a \emph{privacy-preserving federated learning framework for ML applications} in the AMI. 
We consider each smart meter as a federated edge device hosting an ML application that exchanges information with a central aggregator or a data concentrator, periodically. Instead of transferring the raw data sensed by the smart meters, the ML model weights are transferred to the aggregator to preserve privacy. The aggregator processes these parameters to devise a robust ML model that can be substituted at each edge device. We also discuss 
strategies to enhance privacy and improve communication efficiency while sharing the ML model parameters, suited for relatively slow network connections in the AMI. 
We demonstrate the proposed framework on a use case federated ML (FML) application that improves short-term load forecasting (STLF). 
We use a long short-term memory (LSTM) recurrent neural network (RNN) model for STLF. In our architecture, we assume that there is an aggregator connected to a group of smart meters. The aggregator uses the learned model gradients received from the federated smart meters to generate an aggregate, robust RNN model which improves the forecasting accuracy for individual and aggregated STLF. 
Our results indicate that with FML, forecasting accuracy is increased while preserving the data privacy of the end-users.
\end{abstract}
\begin{IEEEkeywords}
Smart Meter Data Analytics, Federated Machine Learning, Short-term load forecasting, user privacy
\end{IEEEkeywords}
\IEEEpeerreviewmaketitle
\section{Introduction}
\label{sec_intro}
Smart meters are being rapidly deployed in household operations with an estimated $132$ million smart meters to be operational in the United States alone, by the end of $2020$~\cite{USDOE2016AMI}. 
Smart meters, with bidirectional communication capability and back-end data management system, make up the advanced metering infrastructure (AMI). The AMI primarily facilitates the demand-side response (DSR) by enabling the active participation of residential consumers. In addition, AMI data is one of the important sources of real-time monitoring data in the smart distribution grid. This has paved the way for \emph{AMI data analytics} and has galvanized a wide range of related applications, such as energy metering, load forecasting, load analysis, load management, electricity theft detection, and power systems monitoring~\cite{wang2019smartMeterreview}. AMI data has also been used to estimate distribution system topologies for enhancing the resiliency of the grid \cite{Peppanen2015AMImodelCal}. 

The recent advances in machine learning (ML) have taken data analytics to the next level, proving to be extremely efficient for modeling complex non-linear interactions. Particularly, the deep architectures of the neural networks in conjunction with enhanced computing capabilities have been successful in solving highly non-linear sequential time-series modeling tasks, such as speech recognition~\cite{Goodfellow-et-al-2016}. The AMI data being essentially sequential time-series has the potential to harness the advancement in ML to develop new applications for \emph{AMI data analytics}.

\subsection{User privacy issues in the AMI}
The smart meter data can reveal certain personal attributes about the consumers, which might be unacceptable from a consumer privacy standpoint. For instance, the individual loads and their schedule of operations can be obtained by decomposing the smart meter data~\cite{Jin2014subgroupTII}. 
In~\cite{Kwac2018tsg}, the authors show that a customer's lifestyle can be categorized by analyzing their household smart meter data.
It can reveal sensitive personal information about a customer, such as availability at home, certain habits, social traits, economic status, etc. 
Therefore, user privacy is a significant concern that needs to be addressed in AMI data analytics applications. 

Most of the privacy preservation techniques in the literature concentrate on the aggregation of the raw smart meter data on local aggregators in an attempt to dilute the identity of the individual customers~\cite{Asghar2017smartMeterPrivacySurvey}. 
The major limitation of this approach is that the aggregator can be compromised to get access to raw data. Moreover, aggregated data can still possess significant pointers, which can be exploited to estimate individual customer behavior. 
Another approach to preserve privacy is to provide confidentiality of data in transit, yet the distribution system operator (DSO) via the aggregators, receives the raw data sensed by the smart meter. Therefore, privacy is an issue if the DSO either wants to use the data for analytics applications or wants to share it with any third-party providers. In the worst case, even data confidentiality is an issue if the aggregator or the DSO are compromised as they contain the raw data in unencrypted form. 

In this paper, we argue that the ML-based applications in the AMI can take the advantage of the black-box nature of ML models to inherently preserve privacy and overcome these limitations. The AMI data analytics applications are generally deployed at the DSO and process the data reported by a group of smart meters. We suggest that the local ML applications can be deployed for each consumer either on a home energy management system (HEMS) device, as a software application on the smart meter, or on an associated embedded device. Considering the progress in Graphical Processing Units (GPUs) and neuromorphic computing, the deployment of ML applications on embedded devices is a reasonable assumption. The ML model weights or parameters are shared with the aggregators (or DSO) alleviating the need to share the raw user consumption data. 
We treat the smart meters as distributed clients hosting efficient local ML applications whose performances can be enhanced by amalgamating the model parameters of all the smart meters in a cluster composed of consumers in a small geographic area, without any violations in user privacy.

\subsection{Contributions}
The {\em key contributions} of this paper are as follows:
{\bf 1)} A generalized privacy-preserving AMI federated ML (AMI-FML) framework for AMI data analytics--the framework is general and can be used as the structural basis for distributed/federated ML applications in the smart grid. {\bf 2)} Strategies to improve communication efficiency and additionally reinforce privacy by minimizing unintended memorization of ML models.
{\bf 3)} A proof-of-concept implementation of the framework using a long short-term memory (LSTM) recurrent neural network for STLF. 
User privacy is better preserved with our AMI-FML framework as we share only the gradient information of the individual smart meter ML models instead of the raw data. Results show the effectiveness and improvement in accuracy due to the use of the federated learning approach.
The rest of this paper is organized as follows. Section~\ref{sec:AMI-Framework} describes the proposed AMI-FML framework. 
Section \ref{sec_methods} presents the proposed AMI-FML application use case for STLF along with brief background on STLF. 
The data description, evaluation criteria, and the framework evaluation with a summary of results and associated discussion are described in Section \ref{sec_experiments}.
Finally, Section \ref{sec_conclusion} presents the concluding remarks.

\section{AMI-FML Framework}
\label{sec:AMI-Framework}
The smart meters in AMI are deployed at the distribution systems level, for each household or small industrial consumers. They measure the consumption of a customer at a specified interval ($15$ minutes, half-hourly, or hourly). We assume that the smart meters are connected to the communication network, capable of receiving pricing information or direct load control commands, and can exchange information with HEMS, which enables optimization of energy usage and demand response. The ML applications can either be installed in the smart meter module, HEMS, or a separate module attached with the smart meter. 

\begin{figure}[!h]
\centering
\includegraphics[width=2.6in]{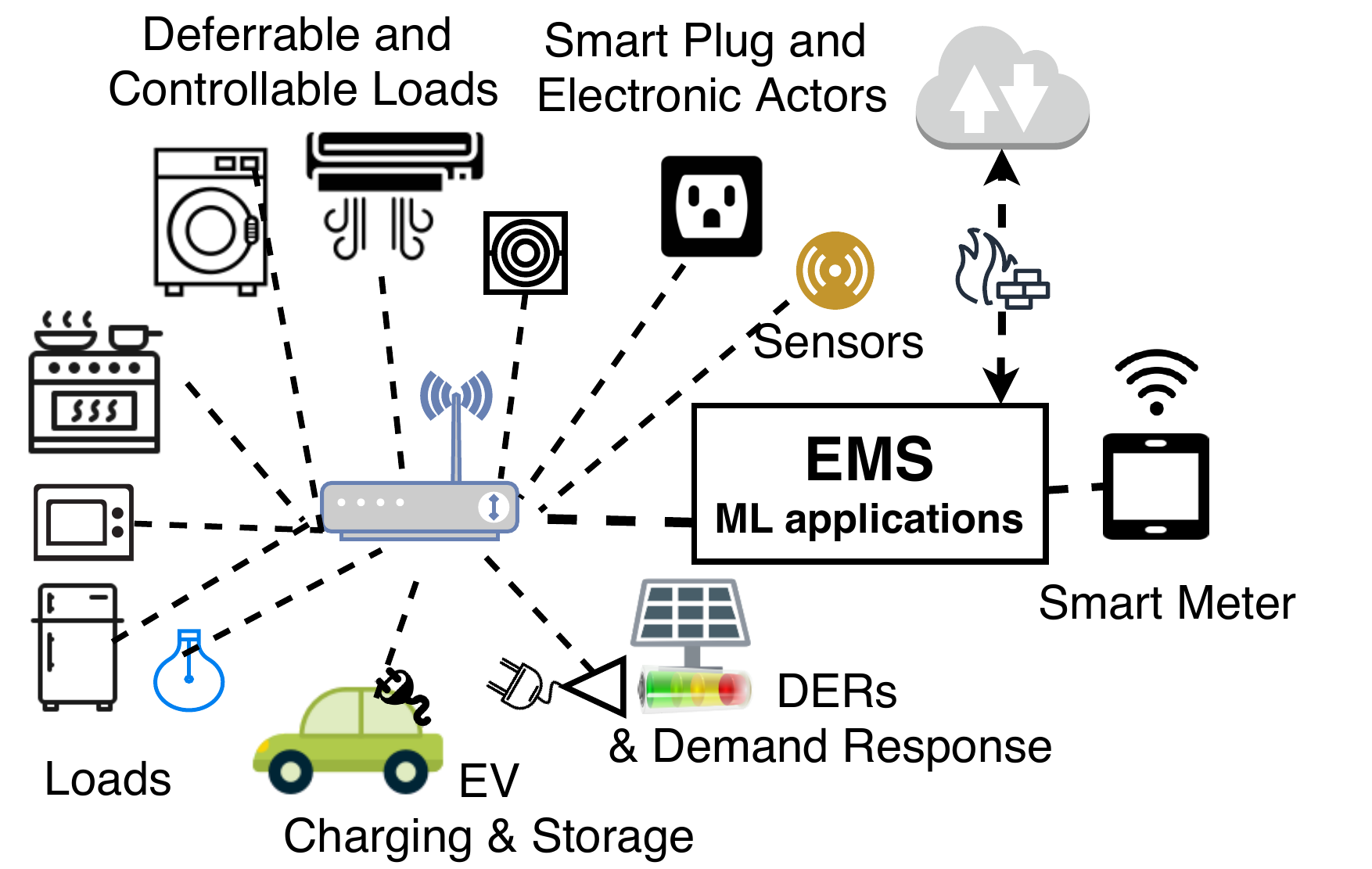}
\caption{Smart meters coupled with the Energy Management Systems}
\vspace{-0.6cm}
\label{fig:ems}
\end{figure}

One of the prominent applications of ML-based AMI data analytics will be to improve the efficiency of the smart distribution grid by keeping an eye on the changing energy markets. The smart meters anchored with HEMS, as illustrated in Fig.~\ref{fig:ems}, can help provide better service to the prosumers. 
Third-party service providers can offer value-added energy services for the HEMS, leveraging smart meter data analytics
~\cite{Kontokosta2013energyMarket}. They could provide commercial energy management solutions in near real-time that can manage loads and distributed generation to reduce energy bills. The third-party demand response (DR) services could rely on ML to achieve 
demand response in coordination with the DR program of the grid operator.
ML-based anomaly detection schemes have been very useful in many systems. 
They can be deployed as smart meter applications to identify anomalous consumption patterns, detect electricity theft, identify power quality disturbances, and monitor equipment health.
The smart meters in a particular geographic region form a cluster and are connected to an aggregator or data center maintained by the DSO. In the case of a third-party ML application, the aggregator can also be hosted on a server connected to the communication network.
Essentially, the federated learning algorithm runs at the aggregator and creates a global ML model by the fusion of local ML models in a cluster. After achieving sufficient generalization of the ML models to minimize errors, the trained ML models are deployed on the smart meters at the edge. 
The fusion of the ML model parameters is carried out at a fixed interval determined by the application requirements. We present a brief overview of the smart meter communication infrastructure in the following section, followed by the details of the FML.


\subsection{Smart Meter Communication Infrastructure}
\label{sec_smart_meter}
\begin{figure}[!h]
\centering
\includegraphics[width=2.4in]{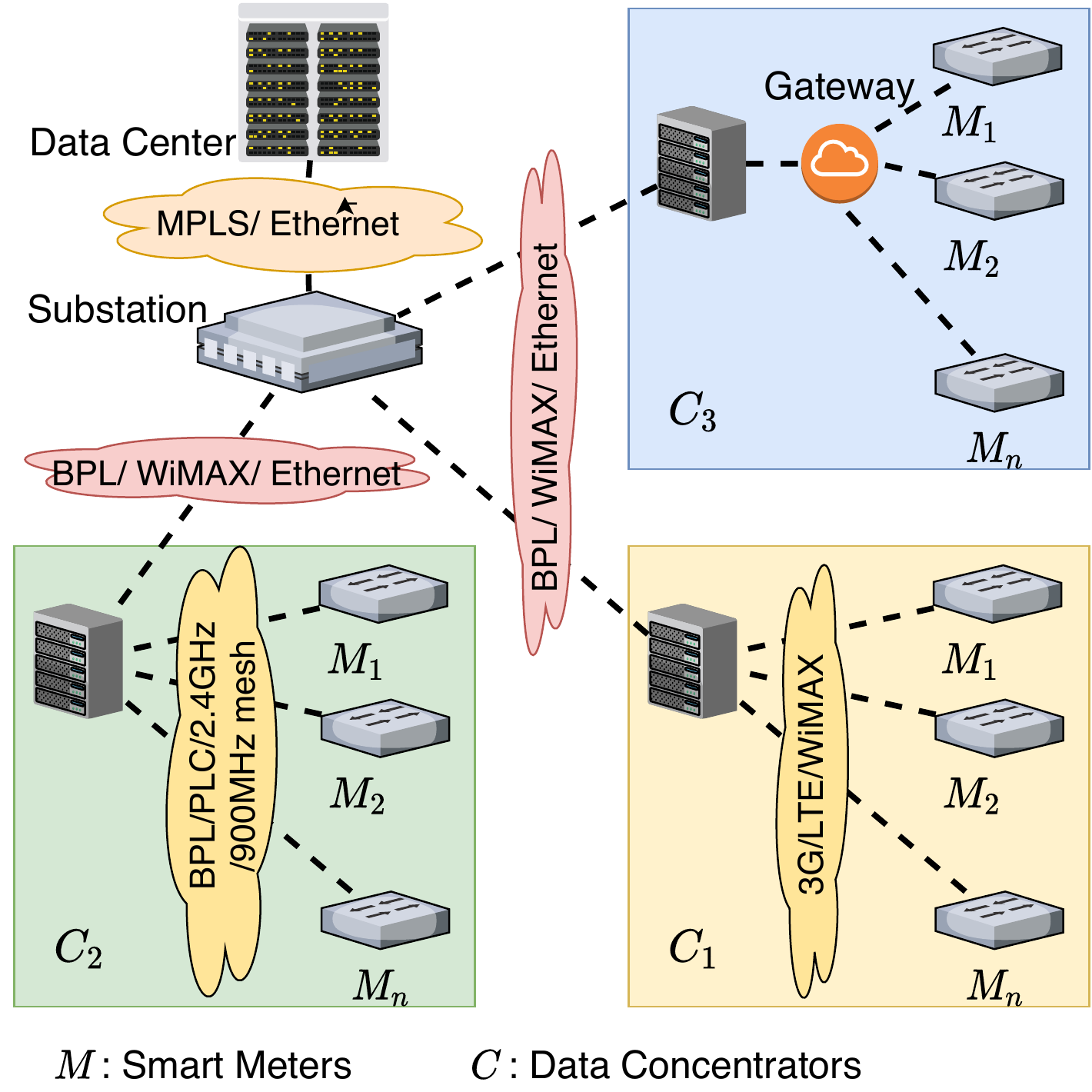}
\vspace{-.3cm}
\caption{Smart meter communication infrastructure. 
}
\label{fig:fed-arch}
\vspace{-.2 cm}
\end{figure}

The smart meter communication infrastructure illustrated in Fig.~\ref{fig:fed-arch}, is mainly built over three types of architectures~\cite{Chren2016smartmetersInfra}. First, the smart meters are directly connected to the data concentrator or aggregator 
via the mobile network (GPRS, CDMA, LTE, 5G), and the responsibilities for data communications are handled by the mobile operator. Second, the smart meters are connected to the data concentrator through Power Line Communications (PLC) or Broadband over Power Lines (BPL) technologies. Third, there can be a gateway to which the smart meters are connected and the gateway can dynamically choose the communication media among cellular, WiFi hot-spots, and PLC/BPL network. The gateway acts as an interface between the data concentrators and the smart meters. The data concentrators are connected to the substation via BPL, WiMAX, or Ethernet.

The ML applications at the consumer side can use any of the communication infrastructures to exchange information with the central aggregators. In the near future, the data communications in the smart meter may happen over the \emph{public} or non-dedicated communication networks, such as the internet. This will further allow the third-party service providers to directly interact with the consumers, through the internet. 


\subsection{Federated machine learning}
\label{sec_federated}
The canonical federated ML (FML) deals with building a global statistical model from the data gathered by a group of remote local devices (smart meters) deployed at the edge. 
The global ML model is created at the aggregator (or DSO) by the fusion of all the local ML model parameters corresponding to each smart meter. The aggregator and the smart meters exchange model parameters through intermediate updates communicated periodically. 

The local ML model weights ($W^{t}_n \in \mathbb{R}$) are updated through gradient descent. 
The number of epochs for learning is fixed so that sufficient generalization is achieved with the inclusion of new training data. 
After a fixed time step the learned weight for all the smart meters is sent to the central server or the aggregator, which hosts the global model. Instead of sending the actual weight values to the aggregator, we send the difference in the weight values as the weights update (also referred to as model update). 
The model update sent by the $n^{th}$ smart meter to the aggregator is expressed as:
\begin{equation}
\label{eq:localwtupdate1}
\Delta W^t_{n} = W^t_{n}-\Bar{W}^{(t-1)}
\end{equation}
Here, $\bar{W}^{t-1}$ represents the global ML model weights at the time $(t-1)$.

The aggregator or the centralized server receives the weight updates $\Delta W^t_{n}$ from all the 
individual smart meters in a set $n \in \mathcal{N}$, at any instant of time $t$. These weights are then combined according to the federated averaging algorithm~\cite{mcmahan2017federated} to create a global model that encapsulates information abstracted by the local models. 

For the experiments in this paper, we use the \emph{FedAvg}~\cite{mcmahan2017federated} algorithm to aggregate the local models, where the parameters are averaged element-wise with weights proportional to the sizes of local datasets. However, the algorithm for model fusion can be chosen based on the application at hand.
Typically, the goal of this federated averaging algorithm is  to minimize the following objective function:
\begin{equation}
\label{eq:objFun}
\min_{W^t} {\sum_{n\in \mathcal{N}} p_n F(W^{t}_n) }
\end{equation}
Here, $F(W^{t}_n)$ is the local objective function or the loss function of the local ML model, for $n^{th}$ smart meter at time $t$. $p_n$ specifies the relative contribution of each local model corresponding to a smart meter. For experiments in this paper, as the training data sizes for each local model are assumed to be the same, the value of $p_n$ is set as $\frac{1}{N}$. Here, $N$ is the total number of smart meters connected to an aggregator. 

We perform the federated averaging according to Eqn.~\eqref{eq:globalwtupdate1}, which is essentially the element-wise mean of the model parameters of all the local models.
\begin{equation}
\label{eq:globalwtupdate1}
\Delta W^t = \sum_{n \in \mathcal{N}} p_n \Delta W^t_{n} 
\end{equation}

The global model parameters are updated as:
\begin{equation}
\label{eq:globalwtupdate2}
\bar{W}^{(t+1)} = \bar{W}^t + \Delta W^t
\end{equation}

\subsection{Communication efficiency}
\label{sec_commEfficiency}
The number of parameters required for the acceptable generalization of an ML model depends on the complexity of the data and the nature of the application. A well generalized deep neural network may have thousands of parameters that may burden the communication channel. The transfer of difference in parameter values results in sharing of a small set of the parameters that have been updated. 
In addition to this, there are two umbrella approaches, structured update and sketched update for reducing communication overhead and additionally strengthen privacy~\cite{konevcny2016federated}. 
We discuss these approaches here. 
\subsubsection{Structured Update}
To reduce the cost of sending $\Delta {W^{t}_n}$ to the aggregator, we tested a \emph{structured update} approach. 
In this approach, the update is restricted to have a predefined structure. 
We created a sparse matrix $Y \in {\{0,1\}}$ of the same dimension as $\Delta {W^{t}_n}$, called the \emph{random mask}. 
We created separate random masks for each smart meter independently in every round of federated learning. 
The sparsity of the random mask is proportional to the number of zeros in the corresponding matrix. 
In our experiments, we randomly selected a fraction of the elements in the sparse matrix to be zero. 
The gradient parameter matrix $\Delta {W^{t}_n}$ is then multiplied element-wise with the random mask, to result in a sparse version of $\Delta {W^{t}_n}$, referred to as $\Delta {\Bar{W}^{t}_n} = \Delta W^{t}_n \circ Y $. Here, $\circ$ represents element-wise multiplication operation. Each smart meter, now sends $\Delta {\Bar{W}^{t}}_n$ to the server instead of $\Delta {W^{t}_n}$.

\subsubsection{Sketched Update}
This is an encoding technique to compress the size of the model parameter vectors. Sub-sampling is one of the easiest ways to achieve this, however, it comes with a cost similar to random masking. A more efficient approach is to use probabilistic quantization, where the parameters/weights are distributed into encoded buckets. All the data in a bucket consolidate their values to that of the value associated with the bucket, with an associated probability.

A one-bit adaptive quantization of a matrix $S$ is expressed as:
\begin{equation}
\label{eq:quanization}
    S(i,j)= 
\begin{cases}
    S_{max}, & \text{with probability } \frac{S(i,j)-S_{min}}{S_{max} - S_{min}}\\
    S_{min}, & \text{with probability } \frac{S_{max}-S(i,j)}{S_{max} - S_{min}}
\end{cases}
\end{equation}
where $S_{max} = \max(S)$ and $S_{min} = \min(S)$

\subsection{Privacy in the FML framework}
\label{sec_privacy}
The ML model gradient parameters ($\Delta W^t_{n}$) provide a sparse and encoded representation of the raw smart meter data and thwart the possibility of deriving the input time-series (raw smart meter data) from just model parameters. 
This inherent feature of FML introduces definitive user privacy~\cite{duchi2014privacy}---the gradient parameters cannot be inverted to generate the raw data.
However, the unintentional memorization of ML models could be a concern~\cite{carlini2018secret}. A small amount of unintended memorization in ML models may occur when the models are either over-trained or not regularized. This means that the gradient parameters could be used to infer either some salient features of the raw data or a partial reconstruction of the input data may be possible, which can undermine privacy. 

In the case of the smart meters, the hidden states of the ML model may be exploited to probe for unique consumption patterns. This is alleviated by either randomly dropping some gradient parameters or approximating their values. We perform these inherently to enhance the communication efficiency (for both \emph{random mask} and \emph{adaptive quantization}). This increases the robustness of the AMI-FML against intended or unintended memorization and the privacy is additionally reinforced.
The global ML model essentially consumes the latent features or information that are \emph{exclusively} relevant for the application at hand, instead of the raw smart meter data. Therefore, this framework can be used as a foundation to develop a range of other FML applications without compromising user privacy utilizing the proposed AMI-FML framework.
\section{FML application for STLF}
\label{sec_methods}
In this section, 
we discuss the proposed AMI-FML framework in the context of a use case: ML application for improving STLF while preserving privacy. First, we present a brief overview of the STLF in~\ref{sec_stlf}. In Section~\ref{sec_lstm}, we briefly present the concepts of LSTM-RNN for STLF. 

\subsection{Short-Term  Load Forecasting (STLF)}
\label{sec_stlf}
Both short-term and long-term load forecasting are critical for power distribution utilities. 
The STLF for individual consumers plays a more important role in future grid planning and operation~\cite{kong2019LSTM}.
This enables short-term energy price forecasts which are of particular interest to power portfolio managers. 
The term \emph{short-term} covers from a few minutes up to a few days ahead.
The accuracy of the STLF impacts the efficiency of the DR mechanisms relying on direct load control (DLC) or locational marginal pricing (LMP). LMP is a driver 
and determines distribution congestion pricing (DCP) in the day-ahead energy market~\cite{Liu2014CongestionDR}.
The patterns of the time-series corresponding to the load consumption at individual household levels and feeder levels are generally distinctive, with the former being a lot more volatile than the latter. 

The intrinsic uncertainties associated with STLF come from external factors, such as weather conditions, variable generation from household DERs, and unexpected changes in the demand. 
For instance, the outside temperature has a profound effect on the heating and cooling system loads. The amount of sunshine and wind speed affects the amount of distributed generation pumped into the grid. 
The charging pattern of the electric vehicles and the battery storage further introduce volatility in load consumption. 

The load consumption patterns are a result of the complex nonlinear interactions among several factors including the ones stated above. It is essential to model these interactions with sufficient faithfulness to predict future consumption, which is extremely challenging. This complexity has increased with the rapid rise in the number of grid-connected DERs. 
Several solutions to address this have been proposed, incorporating advances in signal processing and time series forecasting. Autoregressive Integrated Moving Average (ARIMA) type models have been quite successful in modeling non-linear time-varying processes such as STLF~\cite{Maciejowska2016STLF}. However, most of these statistical models cannot encapsulate the long-term dependencies and have a natural tendency to settle towards the mean values of the past series data.

\subsection{LSTM-RNN for load forecasting}
\label{sec_lstm}
In our problem, the local devices are the smart meters hosting applications with local ML models comprising of the LSTM-RNNs. The goal of our ML application is to accurately forecast future energy consumption. Essentially, this is a sequential time series modeling challenge, and RNNs are the ideal candidate for this task. 
The recurrent connections among neurons in an RNN preserve the hidden state transitions associated with the temporal variations. However, the RNNs are difficult to train and suffer from the problem of either exploding gradient (where the weight updates become excessively large numbers) or vanishing gradients (where the weight updates become insignificantly small numbers), hindering the learning process. Another key limitation of the basic RNNs is their inability to preserve the learned information for longer periods with the increase in time lag. 

\begin{figure}[!h]
\centering
\includegraphics[width=2.4in]{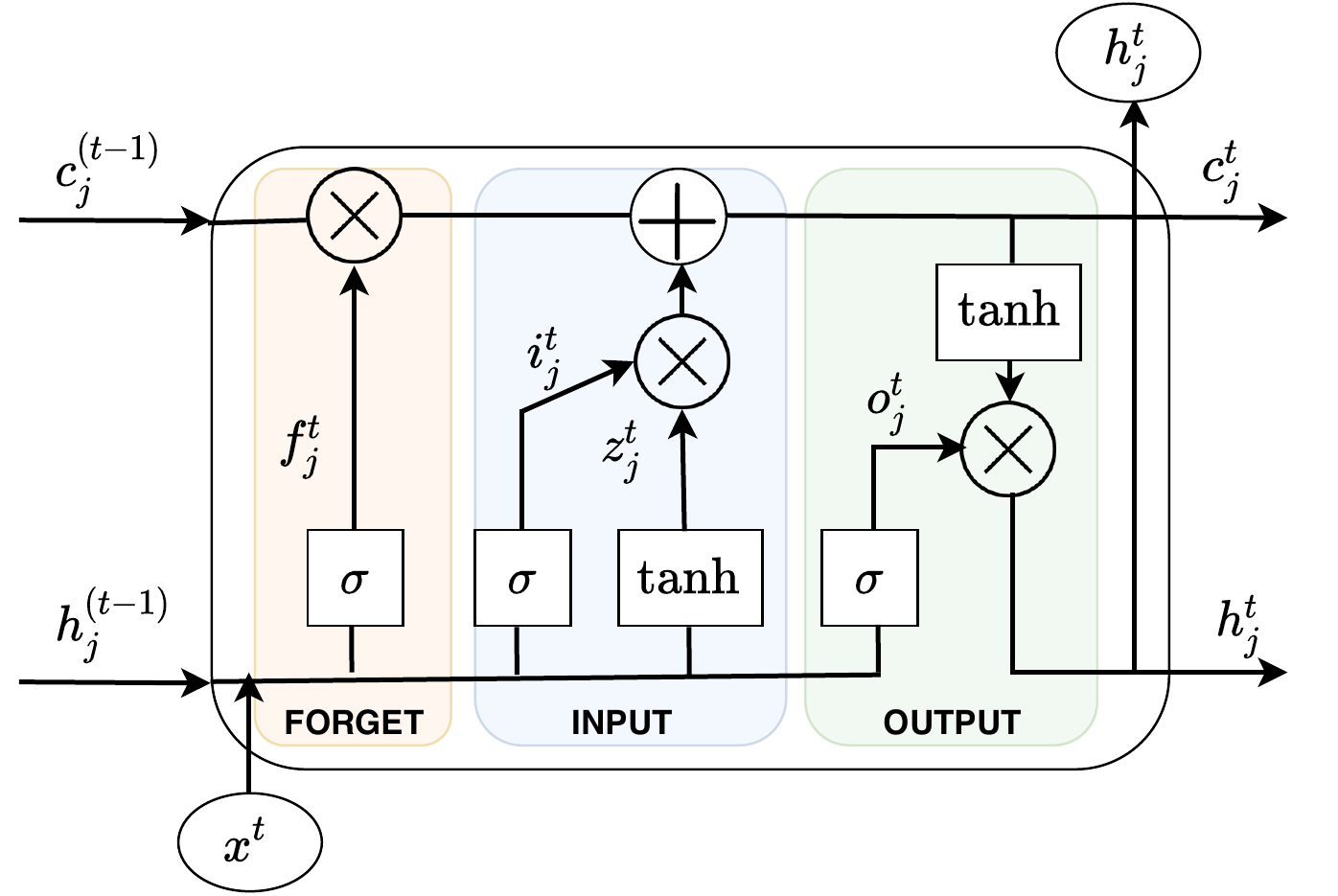}
\vspace{-.4cm}
\caption{Inside an LSTM cell}
\vspace{-.5cm}
\label{fig:lstm-1}
\end{figure}

\begin{figure}[!h]
\centering
\includegraphics[width=2.4in]{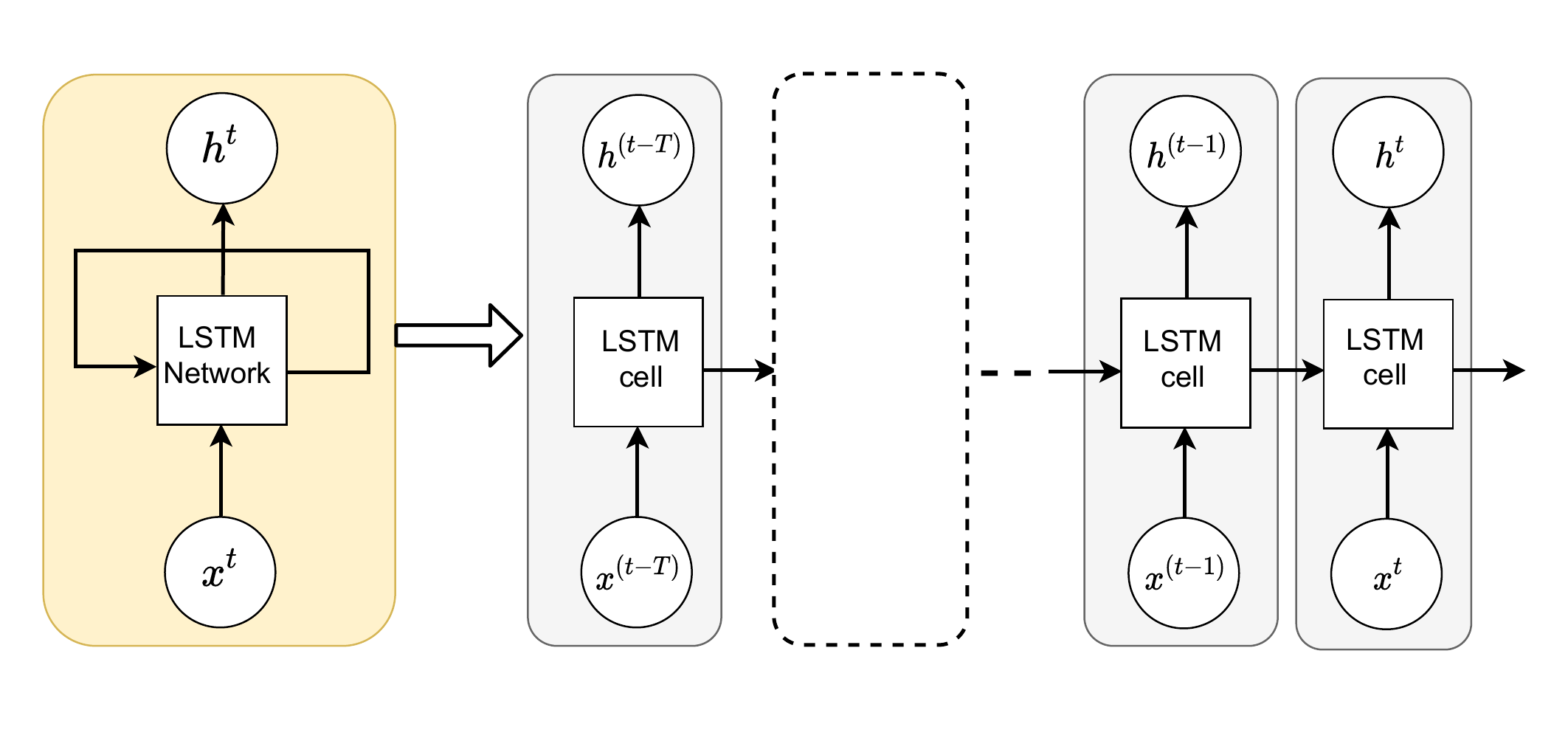}
\vspace{-.5cm}
\caption{Unrolled LSTM network architecture for short-term load forecasting. 
} 
\vspace{-.5cm}
\label{fig:lstm-2}
\end{figure}

The LSTM is a variant of RNN, that alleviates these limitations, paving the way for efficiently modeling sequential data. 
It has demonstrated quite remarkable short-term load forecasting performance for individual customers~\cite{kong2019LSTM}. 
The key elements in an LSTM cell are the memory cell and the gate units (forget, input, output)~\cite{Greff2017LSTM} as depicted in Fig.~\ref{fig:lstm-1}. 
The \emph{memory cell} is responsible for preserving information regarding the temporal state of the neural network and consequently the temporal patterns of the time-series. The gates consist of multiplicative units that control the flow of information, and decide what and how much information should be preserved. 

Let $x^t$ represents the input to $j^{th}$ LSTM cell and $h^t_j$ represents the internal state of the LSTM cell, at a time index $t$.
The \emph{forget gate} in the LSTM has a sigmoidal activation function ($\sigma$), and is responsible to determine what information is irrelevant and should be erased 
from the memory parameter $ c^{(t-1)}_j $. The \emph{input gate} adds new information $ i^t_j $ to the estimated memory state $z^t_j$.  
Terms $i^t_j$  and $z^t_j$ are derived from the input $x^t$ by processing it through a sigmoidal and a $tanh$ activation function, respectively. The \emph{input gate} also adds the sharing parameter vector $h^{t - 1}_j$ to $c^t_j$. The \emph{output gate} determines the new hidden state $h^t_j$ from the memory state parameter $c^t_j$, previous hidden state $h^{t - 1}_j$, and the input $x^t$ by utilizing \textit{sigmoid} and $tanh$ functions.
The set of equations governing the forward pass of the LSTM network can be expressed as~\cite{Greff2017LSTM}:
\begin{align}
\label{eq:lstmWeightUpdate}
f^t = \sigma(W_fx^t+R_fh^{(t-1)}+ P_f\odot c^{(t-1)} + b_f)\\
i^t = \sigma(W_ix^t+R_ih^{(t-1)}+P_i\odot c^{(t-1)}+b_i \\
z^t = \tanh{(W_zx^t+R_zh^{(t-1)}+b_z)} \\
c^t = z^t\odot i^t + c^{(t-1)}\odot f^t\\
o^t = \sigma(W_ox^t+R_oh^{(t-1)}+ P_o\odot c^t + b_o)\\
h^t = \tanh{c^t} \odot o^t
\end{align}
Here, $f^t=[f^t_1 f^t_2 \dots f^t_H]$, $i^t=[i^t_1 i^t_2 \dots i^t_H]$, $z^t=[z^t_1 z^t_2 \dots z^t_H]$, $c^t=[c^t_1 c^t_2 \dots c^t_H]$, and $o^t=[o^t_1 o^t_2 \dots o^t_H]$ are vectors comprising of the \textit{sigmoid} activation outputs of the forget gate, the input gate, $tanh$ activation output of the input gate, cell state, and the \textit{sigmoid} activation output of the output gate respectively.

$W^t_i$, $W^t_z \in \mathbb{R}^{H \times D}$ are matrices consisting of \emph{input gate} weights where $D=1$ is the number of features (one value for each electricity load data point measured by the smart meter) and $H$ is the number of hidden units in the LSTM cell. 
$W^t_f$, $W^t_o \in \mathbb{R}^{H \times D}$ are the forget and output gate weight matrices, respectively. $R^t_z, R^t_i, R^t_f , R^t_o \in \mathbb{R}^{H \times H}$ are the recurrent weights.
Similarly, $P^t_f , P^t_i, P^t_o \in \mathbb{R}^{H}$ are peephole weights and  $b_f^t, b_i^t, b_z^t, b_o^t \in \mathbb{R}^{H} $ are the bias parameters.
The cumulative parameter set of an LSTM network corresponding to the $n^{th}$ smart meter at any time index $t$ consists of all the weights and bias values for that smart meter, which is represented as a multidimensional data frame $W^t_n$~\cite{Greff2017LSTM}.
\begin{equation}
\label{eq:lstmWeights}
W^t_n = \begin{Bmatrix}
W^t_{f} & R^t_{f} & P^t_{f} & b^t_{f} \\
W^t_{i} & R^t_{i} & P^t_{i} & b^t_{i} \\
W^t_{z} & R^t_{z} &  \-- & b^t_{z} \\
W^t_{o} & R^t_{o} & P^t_{o} & b^t_{o}
\end{Bmatrix}
\end{equation}

In Fig.~\ref{fig:lstm-2}, we show a general LSTM network architecture and its unrolled representation. 
The smart meter data $x(t)$ for a given smart-meter is sequentially fed into the LSTM network in the form of an input vector $x(t) = [x^{(t-T)}_D~\dots ~x^{(t-1)}_D~x^{(t)}_D ]$ corresponding to a time instant $t$. Here, $T=48$ is the number of previous time steps (electricity load data points) we want to use to predict the electricity load of the next time step. 
Since the LSTM cells contain \textit{sigmoid} and $tanh$ activation functions which are sensitive to the scale of the data, we normalized our data in the range $[0-1]$. 
After feeding input to the LSTM network, it creates an output of shape $[T, H]$. 
As we used $50$ hidden units for each of our LSTM cells, we have $H=50$. This implies, in the output for each time step there will be $50$ hidden features. In the end, we took $50$ hidden features from the last time step of the LSTM output and passed it through a linear layer to get the final output which is the predicted energy consumption at the time $(t+1)$. 

\section{Evaluation and Results}
\label{sec_experiments}
We evaluated the proposed AMI-FNL framework on the ML based STLF use-case (discussed in Section \ref{sec_methods}). The dataset for the experiments and the performance metrics are described in Sections \ref{sec_data} and \ref{sec_perf}. We describe the experimental setups and associated results in Sections~\ref{sec_STLF_local}-\ref{sec_Comm_eff}. 
\subsection{AMI Data Description}
\label{sec_data}
We used the dataset released by Smart Metering Electricity Customer Behaviour Trials (CBTs) initiated by Commission for Energy Regulation (CER) in Ireland, for the experiments in this paper. From this dataset, we selected a subset of $3600$ smart meters which were deployed for residential consumers. 
The selected dataset consists of electricity load consumed by the consumers from July $14th$, $2009$ to December $31st$, $2010$ ($533$ days), each comprising of $25,584$ data points, with a sampling interval of $30$ minutes. 
We did pre-processing of the raw data to ensure that the bad data (smart meter readings for the days with missing measurements) were removed. 
In our experiments, we have used the first $503$ days of data ($24,144$ data points) for training and the last $30$ days of data ($1440$ data points) for testing. 

We divided the selected dataset into $10$ clusters with $50$ smart meters in each cluster. In real-world applications, number of 
smart meters in a geographic area connected to an aggregater could form a cluster. In case of third party applications, contingent on the internet, clusters can be carefully formed based on geographical or electrical proximity. The number of smart meters in a cluster can be decided based on the efficiency of the communication infrastructure.   
For the STLF, as the reporting intervals of the smart meters are $30$ minutes, we consider a half-hourly forecasting horizon.

\subsection{Performance Metrics}
\label{sec_perf}
We considered {\em two} well-established performance metrics~\cite{Zhang2020STLF} 
to assess the forecasting performance of the proposed method.
Let, $\hat{y}(m)$ represents 
a point forecast (predicted consumption) at a time index $m$, $y(m)$ is the actual consumption, and the total number of point forecast in an experiment is $M$, for an individual smart meter. 

The performance metrics are defined as: \\(a) Normalized Root Mean Square Error (NRMSE)
    \begin{equation}
    NRMSE = \frac{\sqrt{\frac{\sum_{m=1}^M{(y(m)-\hat{y}(m))}^2}{M}}}{max(y)-min(y)}
    \end{equation}
    Here, $max(y)$ and $min(y)$ represent the maximum and minimum value of actual consumption $y$. \\
    (b) Mean Absolute error (MAE)
    \begin{equation}
    MAE = \frac{1}{M}\sum_{m=1}^M{|y(m)-\hat{y}(m)|}
    \end{equation}

\vspace{-.3cm}
\label{sec_results}
\subsection{STLF with local ML model}
\label{sec_STLF_local}
We did not perform any federated averaging in this experiment. The LSTM models were trained on the historical data (of $503$ days) corresponding to that specific consumer. 
We limited the number of training epochs of the LSTM-RNNs to $4$ in order to achieve acceptable generalization without over-fitting. 
After that, the trained models were deployed for forecasting.
In this case, the LSTM models entirely rely on the available local data for training and learning. 
We computed the performance metric values for individual smart meters and averaged them across all the smart meters in a cluster. 
The overall (average) NRMSE, and MAE across the $10$ clusters were: $0.11$ and $0.41$ respectively.
The small value of NRMSE and MAE indicates the efficacy of LSTM networks for STLF. 
The superior forecasting performance of the LSTM can be attributed to its property that preserved very long-term information that relate the internal state transitions and is essential for accurate forecasting.

\begin{figure}[!h]
\centering
\subfloat[]{\includegraphics[width=3.5in, height=1.8in]{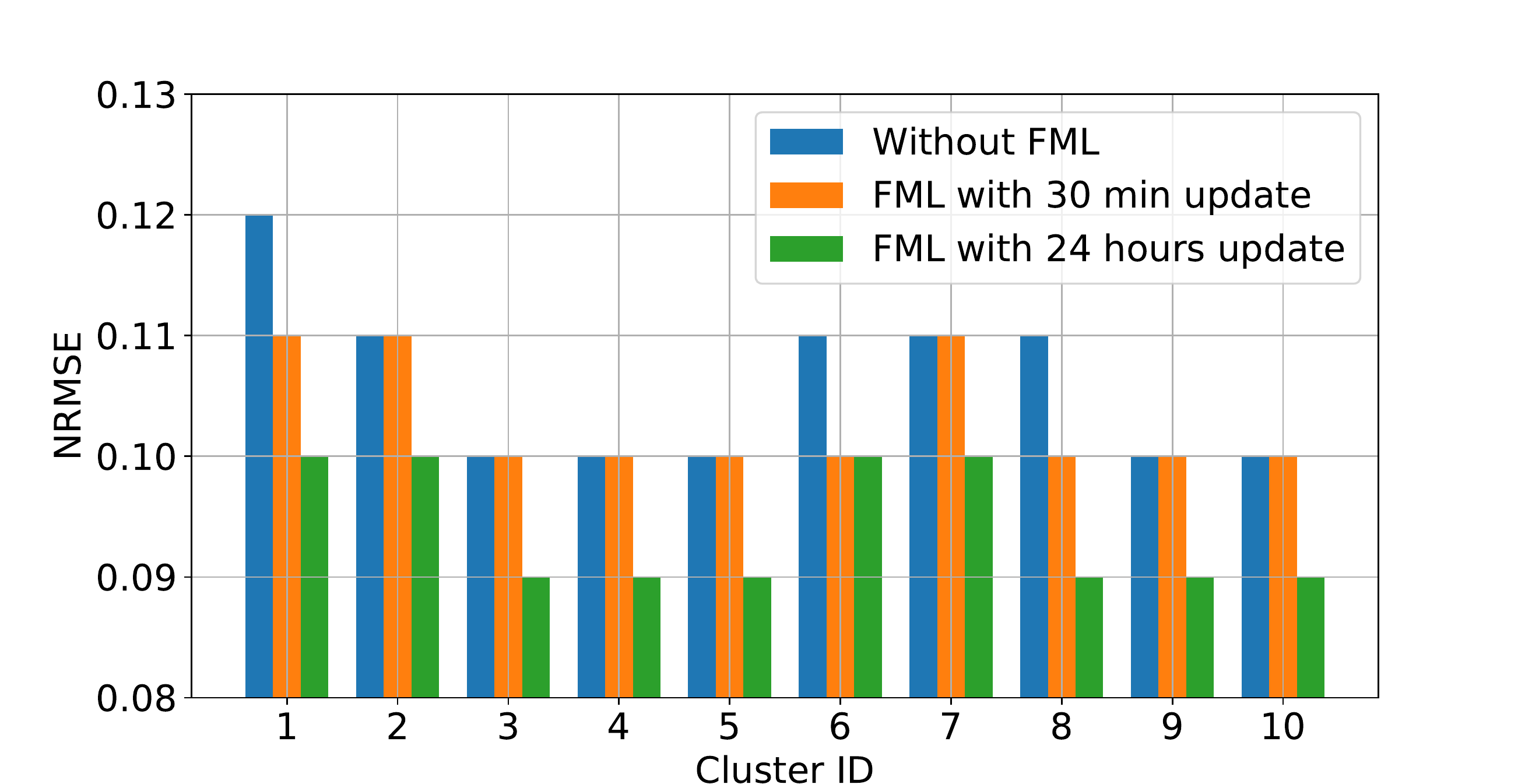}}
\hfill
\vspace{-.05cm}
\subfloat[]{\includegraphics[width=3.5in, height=1.8in]{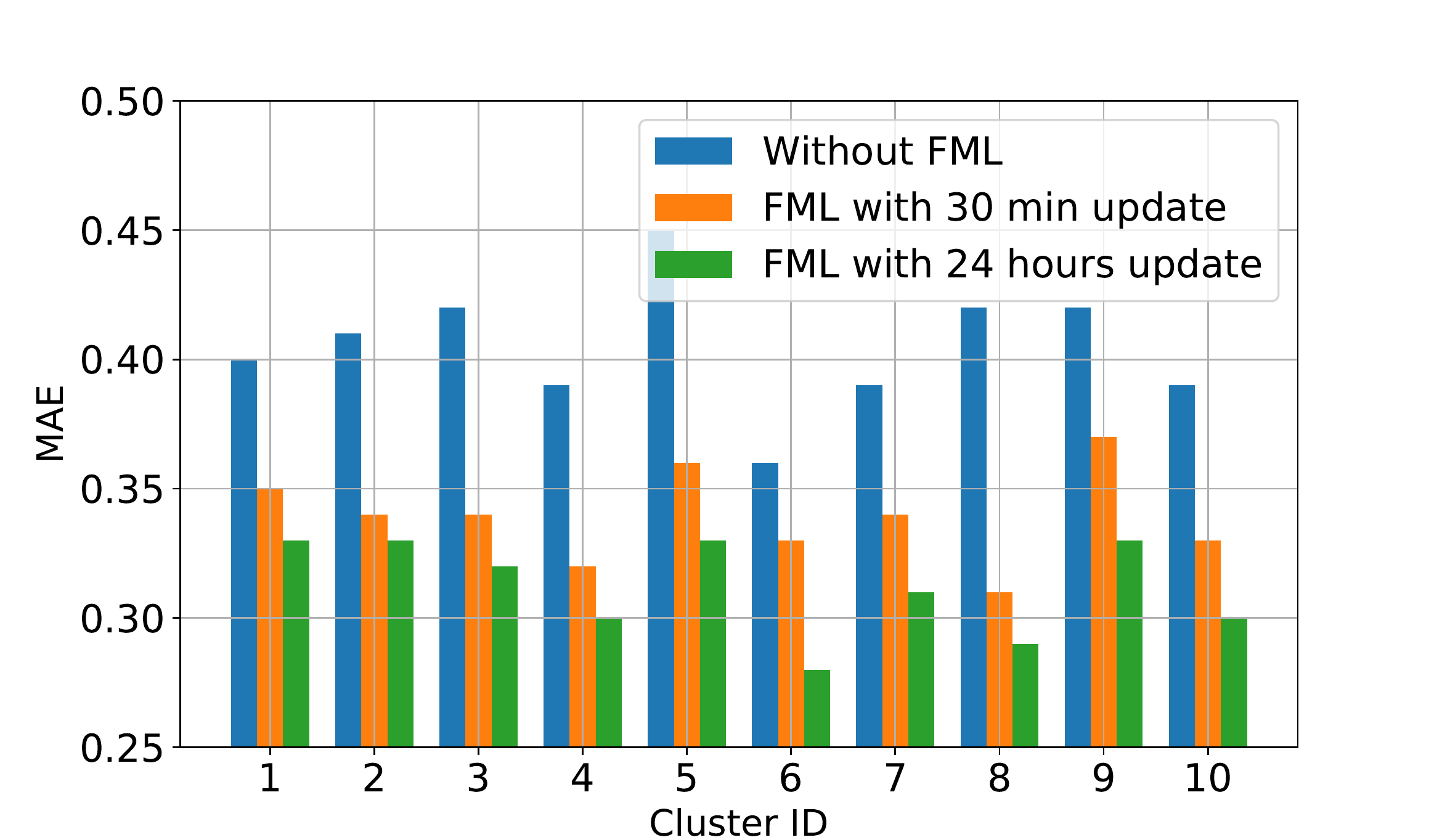}}
\caption{Bar chart showing average (a) NRMSE and (b) MAE for each cluster.} 
\vspace{-.4cm}
\label{fig:errorbar}
\end{figure}

\subsection{STLF with FML model}
\label{sec_STLF_FML}
In this case, we trained the LSTM models in a federated setting by sharing the model weights for each epoch (after traversing through all the time steps in an epoch). 
In the federated ML environment, the model updates are normally shared at specific intervals of time. 
The true advantage of federated ML to improve forecasting accuracy can be harnessed only when we continuously improve the performance of the global model and substitute it in place of the local models at specific intervals. This enables the trained forecasting model to be further trained to accommodate new patterns of consumption after getting deployed at the edge device.

We conducted two sets of experiments related to this. In the first, the local models were shared every $30$ minutes, which was the highest level of granularity that was achievable with our dataset. 
The average NRMSE and MAE averaged across all the clusters were found to be $0.10$, and $0.33$. All these error metrics are smaller compared to the case without federated ML. For all the clusters the average errors decreased or the forecasting performance improved with federated ML. The best-case 
performance improvements (decrease in error) were observed to be $0.11$ for MAE. Although small, these improvements will enhance STLF and impact the transactive energy market.

Then, we conducted another set of experiments where we decreased the granularity to $24$ hours or daily. The motivation behind this is that the consumption patterns are more prominent when we observe the consumption for several days. 
In addition, many utilities give increased emphasis on the day ahead price forecasting to commit energy transactions. 
The average NRMSE and MAE with $24$ hourly federated model updates were observed to be $0.094$ and $0.312$ respectively.
These errors are smaller compared to the case without federated ML, and even smaller compared to the federated ML case with half-hourly updates. 

The superior performance of the daily model update compared to half-hourly updates can be explained by the functional working principle of the LSTM network. The LSTM learns unique features from the transition patterns embedded in the data, which requires the exploration of sequential data of sufficient length/duration. The inclusion of one point to the training sequence may not create adequate information to be abstracted in the LSTM. Whereas, the $24$ hours update may offer a range of motifs for the consumption patterns. This is because certain loads such as the dishwasher follow a daily routine, and are operational for a specific interval of time. 



\subsection{Communication efficiency}
\label{sec_Comm_eff}
\begin{table}[!t]
\caption{Mean Forecasting errors 
with Structured and Sketched update of model parameters 
every 24 hours.}
\label{table_individual_fed30_random}
\centering
\begin{tabular}{c|c|c|c|c}
\hline
& & NRMSE & MAE \\
\hline
\hline
 & 10\% & 0.12 & 0.42 \\
Random Mask & 5\% &  0.11 & 0.38\\
 &2\% & 0.10  & 0.34\\
 \hline
 & 1 bit &  0.12 & 0.38 \\
Quantization & 2 bit &  0.11 & 0.37\\
 & 4 bit & \textbf{0.10} & \textbf{0.32}\\
\hline
\end{tabular}
\vspace{-.5cm}
\end{table}

In this experiment, we evaluated the techniques for improving communication efficiency. 
The forecasting performance of \emph{random mask} based structured update and \emph{adaptive quantization} based sketched updates are summarized in Table~\ref{table_individual_fed30_random}. The performance reductions for random mask with $10\%, 5\%$, and even $2\%$ zeros are significant. However, for sketched update with $4$-bit adaptive quantization, the performance is close to the original case with 24-hour federated ML updates. 
The quantization, in addition to communication overhead reduction ($4$-bit quantization results in $8\times$ compression compared to a $4$ byte float) 
also introduces ancillary privacy features~\cite{konevcny2016federated}. 


\section{Conclusion}
\label{sec_conclusion}
We presented a generalized federated learning based 
paradigm for developing privacy-preserving ML applications (AMI-FML) in the smart grid. We demonstrated techniques to enhance communication efficiency and reinforce privacy in FML. We presented the proof-of-concept through a FML based STLF application supported by the AMI. We used LSTM networks to accurately forecast energy consumption based on the historical smart meter readings. It was demonstrated that the FML improves the forecasting accuracy of the individual smart meters compared to local ML models (without FML), and aids in more efficient HEMS while preserving user privacy. 
In addition to the facilitation of intelligent services by the grid operators, the proposed framework will specifically encourage third-party service providers to roll out value-added services for the benefit of consumers. 

The future direction could be to bring into play the advances in federated ML concepts to improve the performance of smart meter based ML applications in the smart grid. It will be worthwhile to evaluate advanced federated averaging algorithms such as Agnostic Federated Learning, Probabilistic Federated Neural Matching, and matched averaging. Enhancing communication efficiency while embedding differential privacy without reduction in performance is another aspect of potential future work.

\ifCLASSOPTIONcaptionsoff
  \newpage
\fi


\bibliographystyle{IEEEtran}
\bibliography{IEEEabrv,smartgrid.mb}
%
%








\end{document}